\documentclass[letterpaper]{article}
\usepackage{uai2019}
\usepackage[margin=1in]{geometry}

\usepackage{times}
\usepackage[utf8]{inputenc} 
\usepackage[T1]{fontenc}  
\usepackage{hyperref}    
\usepackage{url}      
\usepackage{booktabs}    
\usepackage{amsfonts}    
\usepackage{nicefrac}    
\usepackage{microtype}   

\usepackage{graphicx}
\usepackage{subfigure}
\usepackage{amsmath}
\usepackage{multirow}
\usepackage{algorithm,bm}
\usepackage{algorithmic}

\usepackage{wrapfig}
\usepackage{lipsum}

\newcommand{\R}{\mathbb{R}}

\newcommand{\E}{\mathbb{E}}

\newcommand{\Normal}{\mathcal{N}}

\newcommand{\myvec}{\operatorname{vec}}

\newcommand{\argmax}{\operatorname{argmax}}

\newcommand{\softplus}{\operatorname{softplus}}

\newcommand{\xb}{\mathbf{x}} 
\newcommand{\zb}{\mathbf{z}}

\newcommand{\wb}{\mathbf{w}}

\newcommand{\hnu}{\bm{\nu}}

\usepackage{color}
\definecolor{darkred}{rgb}{0.6, 0, 0}
\definecolor{gray}{RGB}{0.5,0.5,0.5}

\title{Neural Dynamics Discovery via \\Gaussian Process Recurrent Neural Networks}

\author{} 

\author{ {\bf Qi She} \\
Intel Labs China \\
\texttt{qi.she@intel.com}
\And
{\bf Anqi Wu}  \\
Princeton Neuroscience Institute\\
Princeton University\\
\texttt{anqiw@princeton.edu}
}

\begin{document}

\maketitle

\begin{abstract}
Latent dynamics discovery is challenging in extracting complex dynamics from high-dimensional noisy neural data. Many dimensionality reduction methods have been widely adopted to extract low-dimensional, smooth and time-evolving latent trajectories. However, simple state transition structures, linear embedding assumptions, or inflexible inference networks impede the accurate recovery of dynamic portraits. In this paper, we propose a novel latent dynamic model that is capable of capturing nonlinear, non-Markovian, long short-term time-dependent dynamics via recurrent neural networks and tackling complex nonlinear embedding via non-parametric Gaussian process. Due to the complexity and intractability of the model and its inference, we also provide a powerful inference network with bi-directional long short-term memory networks that encode both past and future information into posterior distributions. In the experiment, we show that our model outperforms other state-of-the-art methods in reconstructing insightful latent dynamics from both simulated and experimental neural datasets with either Gaussian or Poisson observations, especially in the low-sample scenario. Our codes and additional materials are available at \texttt{\url{https://github.com/sheqi/GP-RNN_UAI2019}}.
\end{abstract}

\section{INTRODUCTION}
Deciphering interpretable latent $\textit{regularity}$ or $\textit{structure}$ from high-dimensional time series data is a challenging problem for neural data analysis. Many studies and theories in neuroscience posit that high-dimensional neural recordings are noisy observations of some underlying, low-dimensional, and time-varying signal of interest. Thus, robust and powerful statistical methods are needed to identify such latent dynamics, so as to provide insights into latent patterns which govern neural activity both spatially and temporally. A large body of literature has been proposed to learn concise, structured and insightful dynamical portraits from noisy high-dimensional neural recordings~\cite{byron2009gaussian,macke2011empirical,gao2015high,gao2016linear,wu2017gaussian,pandarinath2018inferring,she2018reduced}. These methods can be categorized on the basis of four modeling strategies (``$\star$'' indicates our contributions in these components): 

\paragraph{\textbf{Dynamical model} ($\star$)} Dynamical models describe the evolution of latent process: how future states depend on present and past states. One popular approach assumes that latent variables are governed by a linear dynamical system~\cite{krishnan2017structured,she2018stochastic}, while a second choice models the evolution of latent states with a Gaussian process, relaxing linearity and imposing smoothness over latent states~\cite{lawrence2004gaussian,byron2009gaussian}. However, linear dynamics cannot capture nonlinearities and non-Markov dynamical properties of complex systems; and Gaussian process only considers the pair-wise correlation of time points, instead of considering explicit temporal dynamics. We argue that the proposed dynamical model in this work is able to both capture the complex state transition structures and model the long short-term temporal dynamics efficiently and flexibly. 

\paragraph{\textbf{Mapping function} ($\star$)} Mapping functions reveal how latent states generate noise-free observations. A nonlinear transformation is often ignored when pursuing efficient and tractable algorithms. Most previous methods have assumed a fixed linear or log-linear relationship between latent variables and mean response levels~\cite{macke2011empirical,gao2015high}. In many neuroscience problems, however, the relationship between noise-free observation space and the quantity it encodes can be highly nonlinear. Gao et al., \cite{gao2016linear} have explored a nonlinear embedding function using deep neural networks (DNNs), which requires a large amount of data to train a large set of model parameters and can not propagate $\textit{uncertainty}$ from latent space to observation space. In this paper, we employ a non-parametric Bayesian approach, Gaussian process (GP), to model the nonlinear mapping function from latent space to observation space, which requires much less training data and propagates uncertainties with probabilistic distributions. 

\begin{table*}[t!]
    \centering 
    \label{model} \scalebox{.95}{
        \begin{tabular}{c|c|c|c|c|c}
            \toprule
            Model & Dynamics & Mapping function & Link function & Observation & Inference \\\hline
            PLDS~\cite{macke2011empirical}    & LDS & Linear & exp& Poisson & LP\\ 
            PfLDS~\cite{gao2016linear}    & LDS & NN & exp & Poisson & VI + inference network\\
            GCLDS~\cite{gao2015high} & LDS & Linear & exp & Count & VI \\ 
            LFADS~\cite{pandarinath2018inferring} & RNN & Linear & exp & Poisson & VI + inference network\\ 
            P-GPFA~\cite{nam2015poisson} & GP & Linear & Identity& Poisson & LP or VI \\
            P-GPLVM~\cite{wu2017gaussian} & GP & GP & exp& Poisson & LP\\
            $\textbf{Ours}: \mathbf{GP}$-$\mathbf{RNN}$    & $\textbf{RNN}$ &$ \textbf{GP}$& $\textbf{exp}$ & $\textbf{Poisson/Gaussian}$ & $\textbf{VI + inference network}$\\
            \bottomrule
    \end{tabular} }
    \caption{Comparison of different models. ``PLDS'': Poisson linear dynamical system~\cite{macke2011empirical}; ``PfLDS'': Poisson feed-forward neural network linear dynamical systems ~\cite{gao2016linear}; ``GCLDS'': generalized count linear dynamical systems~\cite{gao2015high}; ``P-GPFA'': Poisson Gaussian process factor analysis~\cite{nam2015poisson}; ``P-GPLVM'': Poisson Gaussian process latent variable model~\cite{wu2017gaussian}; and our method GP-RNN: Gaussian process recurrent neural networks. ``LDS'' denotes Linear Dynamical Systems. ``LP'' and ``VI'' indicate Laplace approximation and variational inference, respectively.}\vspace{-3mm}
\end{table*}

\paragraph{\textbf{Observation model}} Neural responses can be mostly categorized into two types of signals, i.e., continuous voltage data and discrete spikes. For continuous neural responses, people usually use Gaussian distributions as generating distributions. For neural spike trains, a Poisson observation model is commonly considered to characterize stochastic, noisy neural spikes. In this work, we propose models and inference methods for both Gaussian and Poisson responses, but with a focus on the Poisson observation model. Directly modeling Poisson responses with a non-conjugate prior has an intractable solution, especially for complex generative models. In some previous methods, researchers have used a Gaussian approximation for Poisson spike counts through a variance stabilization transformation~\cite{yu2009variance}. In our framework, we apply an effective optimization procedure for the Poisson model. 

\paragraph{\textbf{Inference method} ($\star$)} In our setting, due to the increased complexity of both the dynamical model and the mapping function, we should provide a more powerful inference method for recognizing latent states. Recent work has focused on utilizing variational inference for scalable computation, which takes advantage of both stochastic and distributed optimization~\cite{hoffman2013stochastic}. Additionally, inference networks improve computational efficiency while still keeping rich approximated posterior distributions. One of the choices for inference networks for sequential data is multi-layer perceptrons (MLP)~\cite{christopher2016pattern}. However, it is insufficient to capture the increasing temporal complexity as the dynamic evolves. Recurrent neural networks (RNNs), e.g., long short-term memory (LSTM) and gated recurrent unit (GRU) structures, are well known to capture dynamical structures for sequential data. We utilize RNNs as inference networks for encoding both past and future time information into the posterior distribution of latent states. Specifically, we use two LSTMs for mapping past and future time points jointly into the mean and diagonal covariance functions of the approximated Gaussian distribution. We show empirically that instead of considering only past time information as other recent works~\cite{chung2015recurrent,gregor2015draw}, using both past and future time information can retrieve intrinsic latent structures more accurately. 

Given current limitations in the dynamical model, mapping function, and inference method, we propose a novel method using recurrent neural networks (RNNs) as the dynamical model, Gaussian process (GP) for the nonlinear mapping function, and bi-directional LSTM structure as the inference network. This combination poses a richly distributed internal state representation and flexible nonlinear transition functions due to the representation power of RNNs (e.g., long short-term memory (LSTM) or gated recurrent unit (GRU) structures). Moreover, it shows expressive power for discovering structured latent space by nonlinear embeddings with Gaussian process thanks to its advantage in capturing uncertainty in a non-parametric Bayesian way. In addition, the bi-directional LSTM with increasing model complexity can further enhance inference capability because it summarizes either the past or the future or both at every time step, forming the most effective approximation to the variational posterior of the latent dynamic. Our framework is evaluated on both simulated and real-world neural data with detailed ablation analysis. The promising performance of our method demonstrates that our method is able to: (1) capture better and more insightful nonlinear, non-periodic dynamics from high-dimensional time series; (2) significantly improve prediction performance over baseline methods for noisy neuronal spiking activities; and (3) robustly and efficiently learn the turning curves of underlying complex neural systems from neuronal recording datasets.

Table~\ref{model} summarizes the state-of-the-art methods for extracting latent state space from \textit{high-dimensional spike trains}\footnote{We focus on exploring intrinsic latent structures from spike trains, and the related works mentioned here are to our knowledge the most relevant with this research line. Although some excellent works take advantages of both RNN structures and Gaussian process for either modeling or inference~\cite{frigola2014variational,nguyen2014collaborative,mattos2015recurrent,svensson2016computationally,eleftheriadis2017identification}, they are out of the scope in this work.} by varying different model components discussed above. In a nutshell, our contributions are three-fold comparing to the listed methods: 
\begin{itemize}
\item We propose to capture nonlinear, non-Markovian, long short-term time-dependent dynamics by incorporating recurrent neural networks in the latent variable model. Different from the vanilla RNN, we achieve a stochastic RNN structure by introducing latent variables;
\item We incorporate Gaussian process for learning nonlinear embedding functions, which can achieve better reconstruction performance for the low-sample scenario and provide the posterior distribution with uncertainty instead of point estimation in neural networks. Together with RNN, we provide a GP-RNN model (Gaussian Process Recurrent Neural Network) that is capable of capturing better latent dynamics from complex high-dimensional neural population recordings;
\item We evaluate the efficacy of different inference networks based on LSTM structures for inference and learning, and demonstrate that utilizing the bi-directional LSTM as the inference network can significantly improve model learning.
\end{itemize}

\section{GAUSSIAN PROCESS RECURRENT NEURAL NETWORK (GP-RNN)}
Suppose we have simultaneously recorded spike count data from $N$ neurons. Let $x_{i,t}$ denote the spike count of neuron $i \in \{1,\dots,N\}$ at time $t \in \{1,\dots,T\}$. We aim to discover low-dimensional, time-evolving ($\zb_{t}$ depends on $\zb_{1:t-1}$) latent trajectory $\zb_{t} \in \R^{L}$ ($L\ll N$, and $L$ is the latent dimensionality), which governs the evolution of the high-dimensional neural population $\xb_{t}=[x_{1,t}, x_{2,t}, ..., x_{N,t}] \in \R^{N}$ at time $t$.

\textbf{Recurrent structure latent dynamics}: Let $\zb_{t} \in \R^{L}$ denote a (vector-valued) latent process, which evolves based on a recurrent structure (RNN) to capture the sequential dependence. At each time step $t$, the RNN reads the latent process $\zb_{t-1}$ at the previous time step and updates its hidden state $\hnu_{t} \in \R^{H}$ by:
\begin{align}
\hnu_{t} = \mathbf{RNN}_{\theta}(\zb_{t-1}, \hnu_{t-1}),
\label{rnn_hidden1}
\end{align}
where $\mathbf{RNN}_{\theta}$ is a deterministic nonlinear transition function with parameter $\theta$. $\mathbf{RNN}_{\theta}$ can be implemented via long short-term memory (LSTM) or gated recurrent unit (GRU). It is denoted that the latent process $\zb_{t}$ is modeled as random variables and $\hnu_{t}$ represents hidden states of the RNN model. We model the latent process $\zb_{t}$ by parameterizing a factorization of the joint sequence probability distribution as a product of conditional probabilities such that: 
\begin{align}
p(\zb_{1}, \cdots, \zb_{T}) &= \prod_{t=1}^{T}p(\zb_{t} | \zb_{1}, ..., \zb_{t-1})= \prod_{t=1}^{T}p(\zb_{t} | \zb_{<t})\nonumber\\
p(\zb_{t} | \zb_{<t}) &= p\big(\zb_{t}; g_{\psi}\left(\hnu_{t}\right)\big),\label{rnn_hidden2}
\end{align}
where $g_{\psi}(\cdot)$ is an arbitrary differentiable function parametrized by $\psi$. The function $g_{\psi}(\cdot)$ maps the RNN state $\hnu_{t}$ to the parameter of the distribution of $\zb_{t}$, which is modeled using a feed-forward neural network with $2$ hidden layers as:
\begin{align}
    p\big(\zb_{t}; g_{\psi}\left(\hnu_{t}\right)\big) &= \Normal\big(\mu_{\zb_{t}}, \mathrm{diag}(\sigma^2_{\zb_{t}})\big),\\
    [\mu_{\zb_{t}}, \sigma^2_{\zb_{t}}] &= \mathrm{NN_{2-layer}}(\hnu_{t}).
\end{align}

\textbf{Nonlinear mapping function}: Let $f_{i}:\R^L\rightarrow\R$ denote a nonlinear function mapping from the latent variable
$\zb_t \in \R^{L}$ to the $i$-th element of the observation vector $x_{i,t} \in \R$. $f_i$ is usually referred as the neuronal tuning curve characterizing the firing rate of the neuron as a function of relevant stimulus in neural analysis. We provide a non-parametric Bayesian approach using Gaussian process (GP) as the prior for the mapping function $f_{i}$. Noticing that $f_i$ is a time-invariant function, we can omit the notation for time step $t$ and describe the GP prior as,
\begin{align}
f_{i}(\zb) &\sim \mathcal{GP}(0, k_{z}),\label{GP}\\
k_{z}(\zb,\zb') &= \rho \exp(\frac{-||\zb-\zb'||^{2}_{2}}{2\sigma^{2}})\label{kernel},
\end{align}
where $k_{z}$ is a spatial covariance function over its $L$-dimensional input latent space. Note that the input $\zb$ is a random variable with uncertainty (eq.~\eqref{rnn_hidden2}). Given that the neuronal tuning curve is usually assumed to be smooth, we use the common radial basis function (RBF) or smooth covariance function as eq.~\eqref{kernel}, where $\zb'$ are arbitrary points in latent space, $\rho$ is the marginal variance and $\sigma$ is the length scale. We stack $f_{i}(\zb_{t})$ across $T$ time steps to obtain $\mathbf{f}_{i} \in \R^{T}$. According to the definition of Gaussian process, $\mathbf{f}_{i}$ forms a multivariate normal distribution given latent vectors at all time steps, as 
\begin{align}
\mathbf{f}_{i} | \zb_{1:T} \sim \Normal(0, \mathbf{K}_{z}),
\label{GP1}
\end{align}
with a $T \times T$ covariance matrix $\mathbf{K}_{z}$ generated by evaluating the covariance function $k_{z}$ at all pairs of latent vectors in $\zb_{1:T}$.
Finally, by stacking $\mathbf{f}_{i}$ for $N$ neurons, we form a matrix $\mathbf{F} \in \R^{N \times T}$ with $\mathbf{f}_{i}^\top$ on the $i$-th row. 

\textbf{Observation model}: Real-world time series data is often categorized into real-valued data and count-valued data. For real-valued data, the observation model is usually a Gaussian distribution given the firing rate $f_{i}(\zb_{t})$ and some additive noise $\epsilon\sim\mathcal{N}(0,l)$. Marginalizing out $\epsilon$, we obtain the observation model as
\begin{equation}
x_{i,t} | f_i, \zb_{t} \sim \mathcal{N}\big(f_{i}(\zb_{t}),l\big).
\label{eq:gaussian}
\end{equation}
However the observation following Gaussian distribution is infeasible under count-valued setting. Considering neural spike trains, we assume that the spike rate $\lambda_{i,t} = \exp\big(f_{i}(\zb_{t})\big)$ (non-negative value), and the spike count of neuron $i$ at time $t$ is generated as 
\begin{equation}
x_{i,t} | f_i, \zb_{t} \sim \mathbf{Poisson}\Big(\exp\big(f_{i}(\zb_{t})\big)\Big).
\label{eq:poisson}
\end{equation}

\begin{figure}[t] 
\centering
\includegraphics[width=0.46\textwidth]{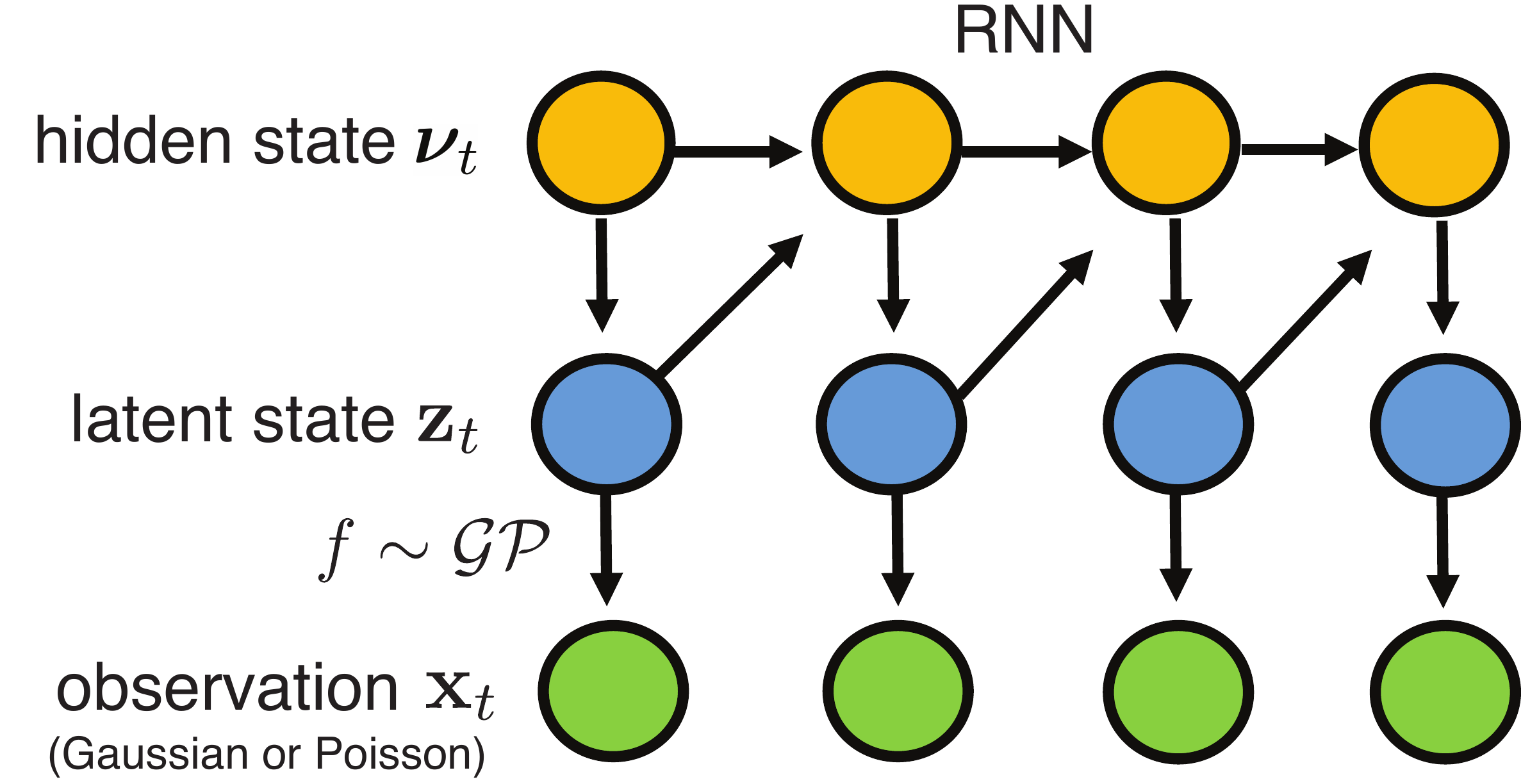} 
\vspace{-3mm}
\caption{The proposed GP-RNN models the dynamics of hidden states $\hnu_t$ (yellow circle) with an RNN structure, and generates latent dynamics $z_{t}$ (blue circle) given hidden states. Both hidden states $\hnu_t$ and latent dynamics $z_{t}$ contribute to $\hnu_{t+1}$. The latent states $z_{t}$ are mapped to observations $x_{t}$ (green circle) via a Gaussian process mapping function $f$.}
\label{model} 
\end{figure} 

In summary, our model uses an RNN structure to capture nonlinearity and long short-term temporal dependence of latent dynamics, while keeping the flexibility of non-parametric Bayesian (GP) in learning nonlinear mapping functions. Finally, we generate Gaussian observations with Gaussian additive noise given spike rates or propagate spike rates via an exponential link function to generate Poisson observations. The graphical model is shown in Fig.~\ref{model}. Denote that RNN structure is not directly applied for latent process $z_{t}$, it is over $z_{t}$'s prior via a neural network mapping (shown in eq.~\eqref{rnn_hidden1} and \eqref{rnn_hidden2}), completely different from a simple RNN for latent states $z_{t}$ as existing works, e.g., LFADS. This modeling strategy, similar to \cite{chung2015recurrent}, establishes stochastic RNN dynamics, which gives a strong and flexible prior over the latent process. $z_{t}$ is propagated with well-calibrated uncertainty via Gaussian process to the firing rate function $f$. The observation $x_{t}$ is generated from $f$ with Gaussian or Poisson noise based on the applications.

\section{INFERENCE FOR GP-RNN}
\textbf{Gaussian response}: When the observation is Gaussian, the tuning curve $f_i$ in eq.~\eqref{eq:gaussian} can be marginalized out due to the conjugacy. Variational Bayes Expectation-Maximization (VBEM) algorithm is adopted for estimating latent states $\zb_{1:T}$ (E-step) and parameters {\small$\Theta=\{\theta, \psi, \rho, \sigma\}$} (M-step). In E-step, we need to characterize the full posterior distribution $p(\zb_{1:T}\vert\xb_{1:T}, \Theta)$, which is intractable. We employ a Gaussian distribution as the variational approximate distribution. Denoting $\bar{\zb} = \myvec(\zb_{1:T})$ and $\bar{\xb} = \myvec(\xb_{1:T})$, we approximate $p(\bar{\zb} \vert\bar{\xb})$ with $q_{\phi}(\bar{\zb}) = \Normal\big(\mu_{\phi}(\bar{\xb}\big), \sigma^{2}_{\phi}(\bar{\xb}))$, whose mean and variance are the outputs of a highly nonlinear function of observation $\bar{\xb}$, and $\phi$ encodes the function parameters. We identify the optimal $\bar{\zb},\Theta$ and $\phi$ by maximizing a variational Bayesian lower bound (also called $``\mathbf{ELBO}''$) as
\begin{small}
\begin{equation}
\!\!\!\mathcal L(\bar{\zb}, \Theta, \phi)    = \E_{q_{\phi}(\bar{\zb})} \big[\log p_{\Theta}(\bar{\zb},\bar{\xb})\big] - \E_{q_{\phi}(\bar{\zb})}\big[\log q_{\phi}(\bar{\zb}) \big]. \label{ELBO}
\end{equation}
\end{small}
The first term in eq.~\eqref{ELBO} represents an energy, encouraging $q_{\phi}(\bar{\zb})$ to focus on the probability mass, $p_{\Theta}(\bar{\zb},\bar{\xb})$. The second term (including the minus sign) represents the entropy of $q_{\phi}(\bar{\zb})$, encouraging it to spread the probability mass thus avoiding concentrating on one point estimate. The entropy term in eq.~\eqref{ELBO} has a closed-form expression:
\begin{equation}
\E_{q_{\phi}(\bar{\zb})}\big[\log q_{\phi}(\bar{\zb}) \big] = -\frac{LT}{2}\big(1+\log(2\pi)\big)-\frac{1}{2}\log|\Sigma|.
\end{equation}
The gradients of eq.~\eqref{ELBO} with respect to $\phi, \Theta$ can be evaluated by sampling directly from $q_{\phi}(\bar{\zb})$, for example, using Monte Carlo integration to obtain noisy estimates of both the ELBO and its gradient~\cite{ranganath2014black,archer2015black}. Score function estimator achieves it by leveraging a property of logarithms to write the gradient as 
\begin{small}
\begin{align}
\nabla \mathcal L(\Theta, \phi) = \frac{1}{S}\sum_{s=1}^{S}\Big[&\nabla \log q_{\phi}(\bar{\zb}_{s})\big(\log p_{\Theta}(\bar{\zb}_{s},\bar{\xb}) \nonumber\\&- \log q_{\phi}(\bar{\zb}_{s})\big)\Big],
 \end{align}
\end{small}
 \begin{table*}[t!]
    \centering
    \begin{tabular}{cccccc}
        \toprule
        Inference Network & Vanilla MF & VAE & r-LSTM &     l-LSTM & bi-LSTM \\ \hline
        Variational Approximation & $q(\zb_{t})$ & $q(\zb_{t}|\xb_{t})$ & $q(\zb_{t}|\xb_{t:T})$ & $q(\zb_{t}|\xb_{1:t})$ & $q(\zb_{t}|\xb_{1:T})$ \\
        \bottomrule
    \end{tabular}
    \caption{Inference networks applied in variational approximation}\vspace{-4mm}
    \label{InferNet}
\end{table*}
which first draws $S$ samples $\{\bar{\zb}_{s}\}_{1}^{S}$ from $q_{\phi}(\bar{\zb})$, and then evaluates the empirical expectation using $\{\bar{\zb}_{s}\}_{1}^{S}$. In general, the approximate gradient using score function estimator exhibits high variance~\cite{ranganath2014black}, and practically we compute the integral with the ``reparameterization trick'' proposed by~\cite{kingma2013auto}. We can parameterize the multivariate normal $\bar{\zb} \sim q(\bar{\zb}|\bar{\xb})$ as
\begin{equation}
\bar{\zb} = \mu_{\phi}(\bar{\xb}) + R_{\phi}(\bar{\xb})\epsilon, \quad \epsilon \sim \Normal(0, I),
\end{equation}
therefore $\zb$ is distributed as a multivariate normal with mean $\mu_{\phi}(\bar{\xb}) $ and covariance $R_{\phi}(\bar{\xb})R_{\phi}(\bar{\xb})^\top$. We finally separate the gradient estimation as
\begin{align}\label{repara}
\nabla_{\Theta}    \mathcal L(\Theta, \phi) &= \E_{q_{\phi}(\bar{\zb})} \big[\nabla_{\Theta} \log p_{\Theta}(\bar{\zb},\bar{\xb})\big], \nonumber\\
\nabla_{\phi}    \mathcal L(\Theta, \phi) &= \E_{\epsilon} \Big[\nabla_{\phi} \log p_{\Theta}\big(\mu_{\phi}(\bar{\xb}) + R_{\phi}(\bar{\xb})\epsilon,\bar{\xb}\big)\Big] \nonumber \\&+ \nabla_{\phi}H_{\phi},
\end{align}
where $H_{\phi} = \E_{q_{\phi}(\bar{\zb})}\left[\log q_{\phi}(\bar{\zb}) \right]$ is the entropy of the variational distribution. Now both gradients can be approximated with Monte-Carlo estimates. 

\textbf{On the choice of the optimal variational distribution:} In eq.~\eqref{ELBO}, we consider the approximated posterior $q_{\phi}(\bar{\zb})$ as a Gaussian, $\Normal\big(\mu_{\phi}(\bar{\xb}), \sigma^{2}_{\phi}(\bar{\xb})\big)$, whose mean and variance are the outputs of a highly nonlinear function of observation $\bar{\xb}$. Here, we consider five structured $q$ distributions by encoding $\bar{\xb}$ in different sequential patterns shown in Table~\ref{InferNet}: (1) vanilla mean field (MF); (2) variational autoencoder (VAE); (3) LSTM conditioned on past observations (l-LSTM); (4) LSTM conditioned on future observations (r-LSTM) and (5) bi-directional LSTM (bi-LSTM) conditioned on both past and future observations. 

\begin{figure}[htbp] 
\centering
\vspace{-3mm}
  \includegraphics[width=0.45\textwidth]{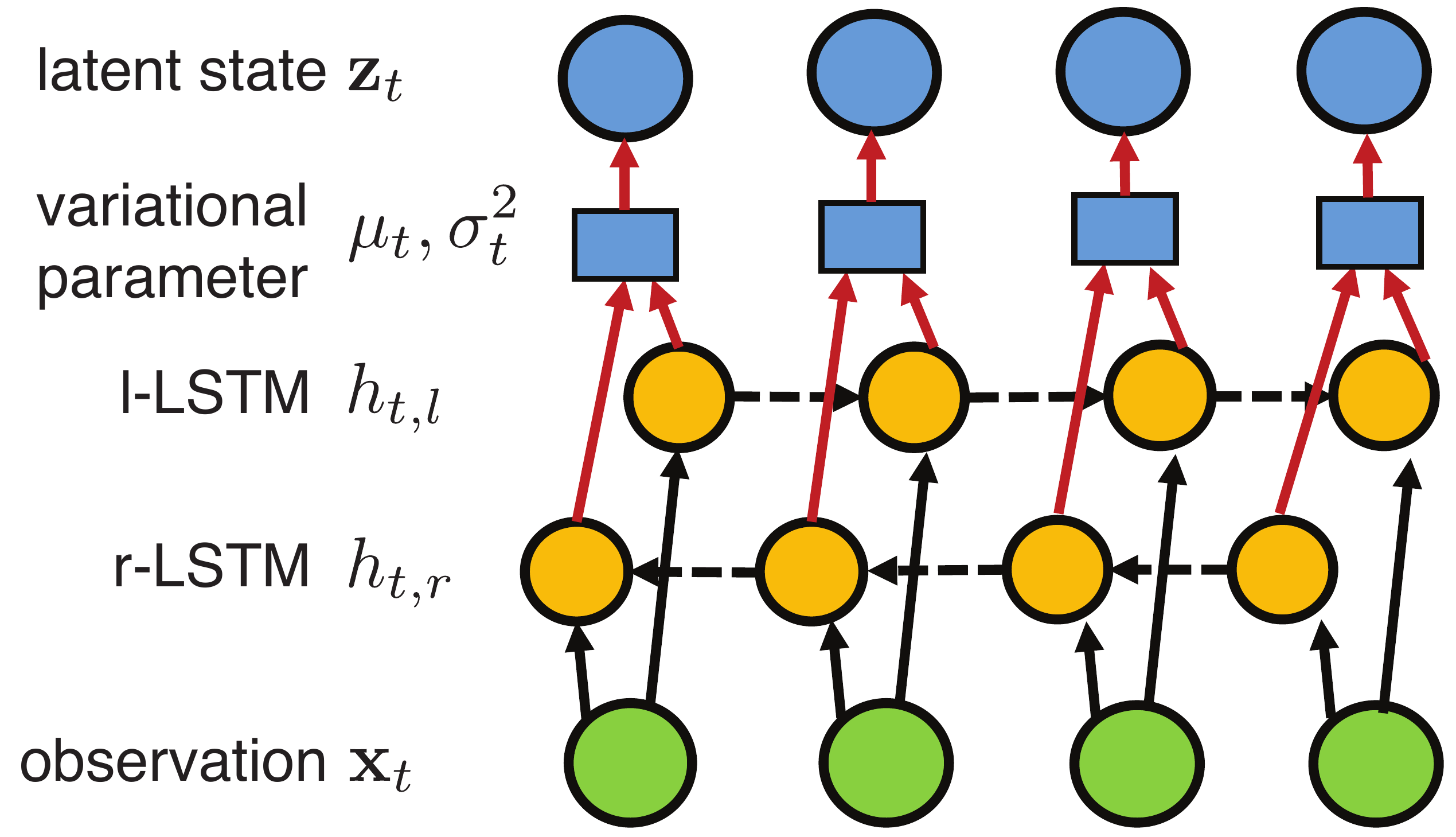} 
\vspace{-3mm}
\caption{Inference network for l-LSTM, r-LSTM and bi-LSTM. Briefly, bi-LSTM is the joint effect of l/r-LSTM. The blue circle denotes latent states $z_{t}$, the blue square shows the variational parameters ($\mu_{t}$, $\sigma^2_{t}$), the yellow circle denotes hidden states of two LSTMs $h_{t,l}$ and $h_{t,r}$, and the green circle represents observation data $x_{t}$.}
\label{bi-LSTM} 
\end{figure}

For l-LSTM and r-LSTM, ``l'' or ``r'' is an abbreviation of ``left'' or ``right'', which considers past or future information. We parametrize mean $\mu_{t}$ and variance $\sigma^{2}_{t}$ for the variational approximated posterior at time step $t$ as a function of the hidden state $h_{t}$, e.g., for l-LSTM, $h_{t,l} = \mathbf{LSTM}(\xb_{1:t})$. We illustrate the l/r/bi-LSTM structure of inference networks in Fig~\ref{bi-LSTM}. Inference network maps observation $\bar{\xb}$ to varational parameters $\mu_t,\sigma^2_t$ of approximate posterior $p(\bar{\zb}|\bar{\xb})$ via LSTM-based structures.
The inference network maps observations to the mean and covariance functions of approximated Gaussian latent states. The parameterization (r-LSTM) can be written as
\begin{align}
\mu_{t,r} &= W_{\mu_{r}} h_{t,r} + b_{\mu_{r}}, \nonumber\\
\sigma^{2}_{t,r} &=\softplus(W_{\sigma^{2}_{r}} h_{t,r} + b_{\sigma^{2}_{r}}).
\end{align}
Similar with the l-LSTM. Here $W$ and $b$ are weights and bias mapping $h_t$ to variational parameters. In bi-LSTM, we use a weighted mean and variance to parameterize the variational posterior as 
\begin{align}
\mu_{t,bi} &= (\mu_{t,r}\sigma^{2}_{t,l}+\mu_{t,l}\sigma^{2}_{t,r})/(\sigma^{2}_{t,l}+\sigma^{2}_{t,r}), \nonumber\\
 \sigma^{2}_{t,bi} &= (\sigma^{2}_{t,l}\sigma^{2}_{t,r})/(\sigma^{2}_{t,l}+\sigma^{2}_{t,r}). 
\label{parameter_mean_variance}
\end{align}
All operations should be performed element-wisely on the corresponding vectors. The Gaussian approximated posterior $q(\zb_{t}|\xb_{1:T}) \sim \Normal(\mu_{t,bi}, \sigma^{2}_{t,bi})$ thus summarizes both the past and future information from observations. 

Algorithm $1$ summarizes the inference method for the Gaussian observation model based on variational inference. 

\begin{algorithm}[htbp!]
            \caption{Inference of GP-RNN-Gaussian}
            \label{alg:A1}
            \begin{algorithmic}
                    \STATE {\bfseries Input:} dataset $\xb_{1:T}$
                    \STATE {\bfseries Output:} latent process $\zb_{1:T}$, model parameters $\Theta=\{\rho,\sigma,\theta,\psi\}$, variational parameter $\phi$
                \REPEAT
                \STATE Evaluate $\mu_{\phi}(\xb_{1:T})$ and $\sigma^2_{\phi}(\xb_{1:T})$ based on~eq.~\eqref{parameter_mean_variance}
                \STATE Sample $\zb_{1:T} \sim \Normal\big(\mu_{\phi}(\xb_{1:T}),\sigma^2_{\phi}(\xb_{1:T})\big)$
                \STATE Evaluate $\mathcal L(\phi, \Theta)$ based on eq.~\eqref{ELBO}
                \STATE Compute $\nabla_{\Theta}    \mathcal L$ and $\nabla_{\phi}    \mathcal L$ based on eq.~\eqref{repara}
                \STATE Update $\Theta$ and $\phi$ using ADAM 
                \UNTIL{$\mathrm{convergence}$}
            \end{algorithmic}
        \end{algorithm}
        
\textbf{Poisson response}: When the observation model is Poisson, the integration over the mapping function $f$ in eq.~\eqref{eq:poisson} is now intractable due to the non-conjugacy between its GP prior and the Poisson data distribution. The interplay between $\zb$ and $f$ involves a highly-nonlinear transformation, which makes inference difficult. Inducing points~\cite{damianou2016variational} and decoupled Laplace approximation~\cite{wu2017gaussian} have been recently introduced to release this dependence and make inference tractable. In this paper, we adapt a straightforward maximum a posteriori (MAP) estimation for training both $\mathbf{F}$ and $\bar{\zb}$, as 
\begin{eqnarray}
\mathbf{F}, \bar{\zb} &=& \argmax_{\mathbf{F}, \bar{\zb}} p(\bar{\xb}, \mathbf{F}, \bar{\zb})\\
&= &\argmax_{\mathbf{F}, \bar{\zb}} p( \bar{\xb}|\mathbf{F}) p(\mathbf{F}| \bar{\zb}, \rho, \sigma)p(\bar{\zb}|\theta,\psi),\nonumber\label{MAP}
\end{eqnarray}
where the joint distribution $p(\bar{\xb}, \mathbf{F}, \bar{\zb})$ of latent variables $\bar{\zb}, \mathbf{F}$ and observations $\bar{\xb}$ of the RHS for eq.~\eqref{MAP} is
\begin{align}\label{joint} 
\!\!\!\! &p(\bar{\xb}, \mathbf{F}, \bar{\zb}) = p( \bar{\xb}|\mathbf{F}) p(\mathbf{F}| \bar{\zb}, \rho, \sigma)p(\bar{\zb}|\theta,\psi) \\
\!\!\!\!&= \prod_{i=1}^{N}\prod_{t=1}^{T}\underbrace{p(x_{i,t}|f_{i,t})}_{\mathbf{Poisson}}\prod_{i=1}^{N}
\underbrace{p(\mathbf{f}_{i} | \bar{\zb}, \rho, \sigma)}_{\mathbf{\mathcal{GP}}}\prod_{t=1}^{T}\underbrace{p_{\theta}(\zb_{t} | \zb_{<t})}_{\mathbf{RNN}}.\nonumber
\end{align}
Eq.~\eqref{joint} is a joint probability with three main components: (1) Poisson spiking (observation model); (2) Gaussian process ($\mathcal{GP}$, nonlinear embedding); and (3) recurrent neural networks ($\mathbf{RNN}$, dynamical model). During the training procedure, we adapt composing inference~\cite{tran2017deep}, fixing $\mathbf{F}$ or $\bar{\zb}$ while optimizing the other in a coordinate ascent manner. More details and the pseudo-algorithm for inference of GP-RNN-Poisson can be found in the supplementary. 

\section{EXPERIMENTS}
\label{Result}
To demonstrate the superiority of GP-RNN in latent dynamics recovery, we compare it against other state-of-the-art methods on both extensive simulated data and a real visual cortex neural dataset. 

\subsection{Recovery of Lorenz Dynamics}
First we recover the well-known Lorenz dynamics in a nonlinear system. The Lorenz system describes a two dimensional flow of fluids with $z_{1,2,3}$ as latent states:
\begin{equation}\label{eq:ld}
\frac{dz_1}{dt} = \sigma(z_2 - z_1), \frac{dz_2}{dt} = z_1(\rho - z_3) - z_2, \frac{dz_3}{dt} = z_1y - \beta z_3.
\end{equation}
This system has chaotic solutions (for certain parameter values) that revolve around the so-called Lorenz attractor. Lorenz uses the values $\sigma =10$ , $\beta =8/3$ and $\rho =28$, exhibiting a chaotic behavior, which generates a nonlinear, non-periodic, and three-dimensional complex system. It has been utilized for testing latent structure discovery in recent works~\cite{zhao2017variational,linderman2017bayesian,pandarinath2018inferring}.

\begin{table*}[t!]
    \centering \scalebox{1.0}{
        \begin{tabular}{c|ccccc|ccccc}
            \hline
            \multirow{2}{*}{Gaussian} & \multicolumn{5}{c|}{AR1-GPLVM}              & \multicolumn{5}{c}{\underline{GP-RNN}}    \\ \cline{2-11} 
            & MF & VAE & r-LSTM & l-LSTM & bi-LSTM & MF & VAE & r-LSTM & l-LSTM & bi-LSTM \\ \hline
            linear & 4.12 & 4.10 & 4.01 & 3.27 & \underline{\textbf{1.64 }} & 2.17 & 2.17 &  1.98  &   1.54  & \underline{\textbf{0.96 }}  \\ \hline
            tanh & 3.20  & 3.22  & 3.01 & 2.46 & \underline{\textbf{1.17 }}  & 2.01 & 2.01 & 1.83  &   1.41  & \underline{\textbf{0.78 }}  \\ \hline
            sine  & 3.12  & 3.12 & 2.74 & 2.33 & \underline{\textbf{1.02 }} & 1.81 & 1.78 &  1.34   &  1.12  & \underline{\textbf{0.56}}  \\ \hline
    \end{tabular} }
    \caption{Inference network and dynamical model analysis. Root mean square error ({\bf{RMSE}}, $10^{-2}$) of latent trajectories reconstructed from various simulated models are presented. We compare two latent dynamical models: first-order autoregressive (AR1) and recurrent neural network (e.g., LSTM), three mapping functions: linear, tanh and sine, and five variational approximations listed in Table~\ref{InferNet}. The observations are Gaussian responses with $50$ observational dimensions and $200$ time points. Underlined and bold fonts indicate best performance. Results with standard errors (ste) can be found in the supplementary.} 
    \label{simulation0}
\end{table*}

\begin{table*}[t!]
    \centering \scalebox{1.0}{
        \begin{tabular}{c|ccccc|ccccc}
            \hline
            \multirow{2}{*}{Poisson} & \multicolumn{5}{c|}{AR1-GPLVM}              & \multicolumn{5}{c}{\underline{GP-RNN}}    \\ \cline{2-11} 
            & MF & VAE & r-LSTM & l-LSTM & bi-LSTM & MF & VAE & r-LSTM & l-LSTM & bi-LSTM \\ \hline
            linear & 6.34 & 6.34 & 6.02 & 5.71 & \underline{\textbf{3.67}} & 6.01 & 6.01 & 5.94 &  5.71 & \underline{\textbf{3.10}}  \\ \hline
            tanh & 3.22   & 3.21 & 3.01 & 2.84 & \underline{\textbf{1.57}}  & 3.09 & 3.11 & 2.98 &  2.54  &  \underline{\textbf{1.21}}   \\ \hline
            sine  & 2.80  & 2.79 & 2.77  & 2.51 & \underline{\textbf{1.49}} & 2.67  & 2.67 & 2.43 &  2.33 &  \underline{\textbf{1.14}}   \\ \hline
    \end{tabular}}
    \caption{Root mean square error ({\bf{RMSE}}, $10^{-2}$) of latent trajectories reconstructed from Poisson responses in test datasets. Underlined and bold fonts highlight best performance. Results with standard errors (ste) can be found in the supplementary.}
    \label{simulation1}
\end{table*}

We simulated a three-dimensional latent dynamic using Lorenz system as in eq.~\eqref{eq:ld}, and then apply three different mapping functions for simulations: $\xb_{t} = \wb^\top \zb_{t} + \Phi + \eta$; $\xb_{t} = \tanh(\wb^\top \zb_{t} + \Phi) + \eta$; and $\xb_{t} = \sin(\wb^\top \zb_{t} + \Phi) + \eta$. Note that the oscillatory response of sine wave is well-known as the properties of grid cells~\cite{gao2016linear}). Thus, we generate simulated data with nonlinear dynamics and linear/nonlinear mapping functions. Gaussian response is the Gaussian noise corrupted version of $\xb_t$; Poisson spike trains are generated from a Poisson distribution with $\mbox{exp}(\xb_t)$ as the spike rate. 

In our simulation, the latent dimension is $3$ and the number of neurons is $50$, thus $\zb_{t} \in \R^{3}$ and $\wb \in \R^{3\times50}$. We randomly generate weights $\wb$ and bias $\Phi$ uniformly from region [0, 1.0], and the noise $\eta$ is drawn from $\Normal(0, I)$. We test the ability of each method to infer the latent dynamics of the Lorenz system (i.e., the values of the three dynamic variables) from Gaussian and Poisson responses, respectively. Models are compared in three aspects: inference network, dynamical model and mapping function. 

\textbf{Analysis of inference network and dynamical model:} 
Table~\ref{simulation0} and \ref{simulation1} show performance of variational approximation techniques applied to both P-GPLVM with AR1 kernel (AR1-GPLVM) and GP-RNN models on Gaussian and Poisson response data respectively. P-GPLVM with AR1 kernel is mathematically equal to the GPLVM model with LDS when the linear mapping matrix in LDS is full-rank. Therefore we are essentially comparing between LDS and RNN for dynamic modeling. In general, GP-RNN outperforms AR1-GPLVM via capturing complex dynamics of nonlinear systems with powerful RNN representations.

bi-LSTM inference networks render best results due to its consideration of both past and future information. Meanwhile, l-LSTM demonstrates the importance of past dependence with better results than r-LSTM. Overall, LSTM-style inference networks have more promising results than models considering current observations only (e.g., MF and VAE). 

Moreover, the inference network of VAE is not a much more expressive variational model for approximating posterior distribution compared with vanilla mean field methods. With only current time points, both of them have similar inference power (as shown in Table~\ref{simulation0} and \ref{simulation1} columns of ``MF'' and ``VAE''). VAE only has global parameters of the neural network for mapping data points to posteriors, while vanilla MF has local parameters for each data point. VAE can be scaled to large-scale datasets, but the performance is upper-bounded by vanilla MF~\cite{tran2017deep}.

\textbf{Analysis of mapping function:} 
\begin{table}[t!]
    \centering \scalebox{1}{
    \begin{tabular}{c|cc|cc|cl}
        \hline
        \multirow{2}{*}{\# Data} & \multicolumn{2}{c|}{linear} & \multicolumn{2}{c|}{tanh}         & \multicolumn{2}{c}{sine}         \\ \cline{2-7} 
        & GP          & NN  & GP         & NN         & GP         & NN         \\ \hline
        N = 50             & \underline{\textbf{2.51}} & 3.88 & \underline{\textbf{1.45}} & 2.75        & \underline{\textbf{1.97}} & 3.43        \\ \hline
        N = 100             & \underline{\textbf{1.27}} & 1.65 & \underline{\textbf{1.15}} & 1.45        & \underline{\textbf{1.03}} & 1.31        \\ \hline
        N = 200             & \underline{\textbf{0.96}} & 1.29 & \underline{\textbf{0.78}} & 1.22        & \underline{\textbf{0.56}} & 0.70        \\ \hline
        N = 500             & \underline{\textbf{0.34}} & 0.35 & \underline{\textbf{0.26}} & \underline{\textbf{0.26}} & \underline{\textbf{0.12}} & \underline{\textbf{0.12}} \\ \hline
    \end{tabular} }
    \caption{Mapping function analysis. {\bf{RMSE}} ($10^{-2}$) of latent trajectory reconstruction using Gaussian process (GP-RNN) and neural network (NN-RNN) mapping functions are shown. Both of them are combined with an RNN dynamical model component. We simulate $50$ trials and present averaged {\bf{RMSE}} results across all trials. Linear, tanh and sine mapping functions are used to generate the data. ``$N$'' indicates the number of data points for training in each trial, and {\bf{RMSE}} is the result of subsequent $50$ time points for testing. Results with standard errors (ste) can be found in the supplementary.}
    \vspace{-5mm}
    \label{NN_GP_Comparison}
\end{table}
\begin{table*}[t!]
    \centering \scalebox{1.0}{
    \begin{tabular}{c|ccccccc}
        \toprule
        Dimension & PLDS & GCLDS & PfLDS & P-GPFA & P-GPLVM & GP-RNN \\\hline
        $z_{1}$ & 0.641 & 0.435 & 0.698 & 0.733 & 0.784& \underline{\textbf{0.869}} \\ \hline
        $z_{2}$ & 0.547 & 0.364 & 0.659& 0.720& 0.785& \underline{\textbf{0.873}}\\ \hline
        $z_{3}$ & 0.903 & 0.755 & 0.797 & 0.960 & 0.966& \underline{\textbf{0.971}}\\
        \bottomrule
    \end{tabular} }
    \caption{$R^{2}$ (best possible score is $1.0$) values of our method and other state-of-the-art methods for the prediction of Lorenz-based spike trains. The included methods are Poisson linear dynamical system (PLDS~\cite{macke2011empirical}), generalized count linear dynamical system (GCLDS~\cite{gao2015high}), Poisson feed-forward neural network linear dynamical system (PfLDS~\cite{gao2016linear}), and Poisson-Gaussian process latent variable model(P-GPLVM~\cite{wu2017gaussian}). GP-RNN recovers more variance of the latent Lorenz dynamics, as measured by $R^{2}$ between the linearly transformed estimation of each model and the true Lorenz dynamics. Results with standard errors (ste) can be found in the supplementary.}
    \label{relatedwork}
\end{table*}

Table~\ref{NN_GP_Comparison} shows the comparison between a neural network and a Gaussian process as the nonlinear mapping functions. The dynamical model is RNN, and the true mapping functions include linear, tanh, and sine functions. The number of data points for training ($N$) are $50$, $100$, $200$ and $500$. The subsequent $50$ time points following the training time points are used for testing the accuracy of reconstructions of latent trajectories. In Table~\ref{NN_GP_Comparison}, We can tell that a Gaussian process provides a superior mapping function for smaller datasets for training (columns of ``GP'' and ``NN''). When we have more time points, the prediction performance of a neural network mapping is comparable with a Gaussian process (rows of $N=200$ and $500$). Bigger datasets can help to learn complex Lorenz dynamics, and meanwhile, prevent the overfitting problem in neural network models. Smaller datasets may affect latent dynamics recovery but a Gaussian process mapping enhances nonlinear embedding recovery via keeping the local constraints. 

\textbf{Comparison with state-of-the-art methods: } 

Consistent with results reported in state-of-the-art methods, we compare $R^{2}$ values for latent trajectory reconstruction of our GP-RNN method against others as shown in Table~\ref{relatedwork}. The inference network of our model is bi-LSTM since the simulated results shown above demonstrate its stronger power in model fitting. Note that we use the Poisson model and compare it with recently developed models for analyzing spike trains. For each dimension of Lorenz dynamics, GP-RNN significantly outperforms baseline methods, e.g., $10.8\%$ ($z_{1}$), $11.2\%$ ($z_{2}$) and $0.5\%$ ($z_{3}$) increment of $R^{2}$ values compared with the second best model P-GPLVM. We have also found several excellent works combining RNN structures with Gaussian process for either modeling or inference~\cite{frigola2014variational,mattos2015recurrent,svensson2016computationally,eleftheriadis2017identification}, but note that they are not in the research line of exploring latent intrinsic structures of high-dimensional real or count-valued data as stated in our work. The methods we compared in our paper (e.g., PLDS~\cite{macke2011empirical}, GCLDS~\cite{gao2015high}, PfLDS~\cite{gao2016linear}, P-GPFA~\cite{nam2015poisson}, and P-GPLVM~\cite{wu2017gaussian}) are to our knowledge recently proposed methods analyzing the same problems and can be more worthwhile being compared.


\begin{figure*}[t!]
    \begin{minipage}[t]{1\textwidth}
        \centering
        \includegraphics[width=0.95\textwidth]{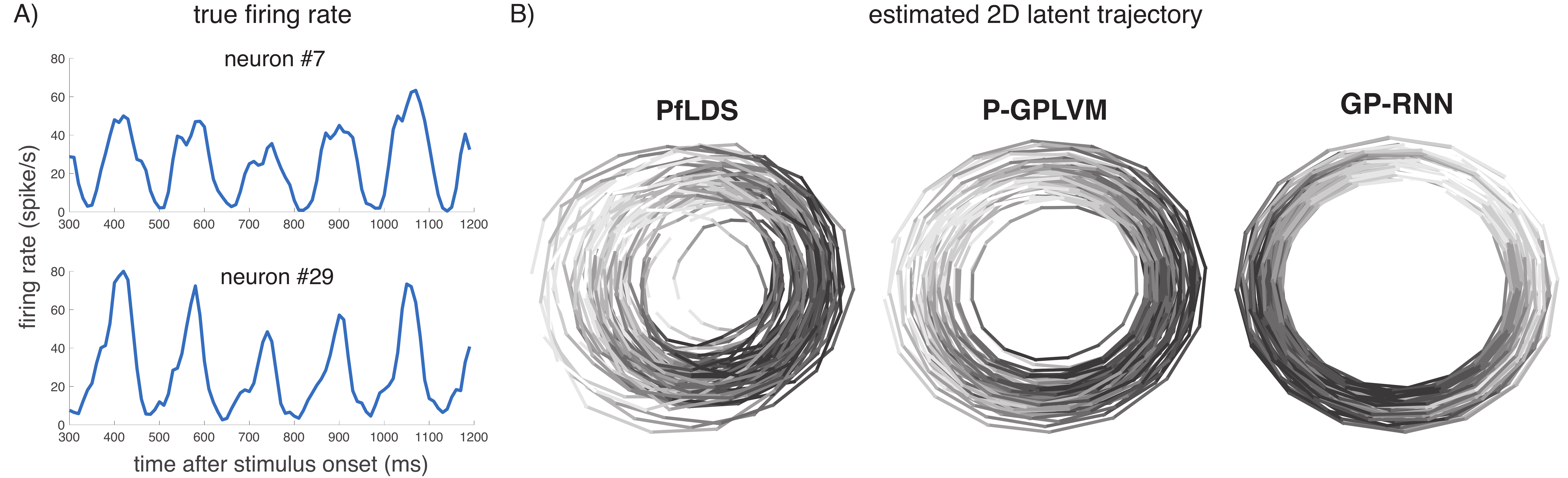}
    \end{minipage}
    \vspace{-5mm}
    \caption{A) True firing rates for $2$ example neurons for orientation $0^\circ$ averaged across 30 trials. We can tell there exists clear periodicity in the firing rate time series given the sinusoidal grating stimulus. B) $2$-dimensional latent trajectories of $10$ out of $30$ trials using PfLDS, P-GPLVM and GP-RNN. Color denotes the phase of the grating stimulus implied in (A). Each circle corresponds to a period of latent dynamics $\zb_{1:T}$ (T = 90) inferred by the models. Each trial is estimated from 63-neuronal spike trains. The latent embedding is smoother and more structured when applying GP-RNN, which is interpretable since the stimulus is sinusoidal for each orientation across time. We can tell that the phase of latent dynamics inferred by GP-RNN is better locked to the phase of the stimulus.} 
    \label{latentstructure}
\end{figure*}

\subsection{Application to Macaque V1 Neural Data}
We apply GP-RNN to the neurons recorded from the primary visual cortex of a macaque~\cite{graf2011decoding}. Data was obtained when the monkey was watching sinusoidal grating driftings with $72$ orientations ($0^\circ$, $5^\circ$, $10^\circ$, $\cdots$, $355^\circ$), and had $50$ repeated trials for each orientation. Following \cite{gao2016linear}, we consider $63$ well-behaved neurons based on their tuning curves, and bin $900$ ms spiking activity with window size $\Delta t = 10$ ms, resulting in $90$ time points for each trial.

We take orientation $0^\circ$ as an example for visualizing $2$-dimensional (2D) latent trajectory estimation. The other orientations exhibit similar patterns. The true firing rates of two example neurons are presented in Fig.~\ref{latentstructure} (A), which exhibit clear periodic patterns locked to the phase of the sinusoidal grating stimulus. In order to get latent dynamics estimation, we fit our model with randomly selected $30$ repeated trials, which are used to learn RNN dynamics parameters and GP hyperparameters shared across all trials, and trial-dependent latent dynamics. We also apply PfLDS and P-GPLVM to the same data. For better visualization purpose,  Fig.~\ref{latentstructure}(B) shows the results of $10$ best trials, which are selected with 10 smallest variances from the mean trajectory within each model. PfLDS has a worse performance compared with the other two methods. Different from the result shown in~\cite{gao2016linear}, we report the result of $30$ trials for training instead of $120$. Benefiting from the non-parametric Bayes (Gaussian process), in such a small-data scenario, GP-RNN extracts much more clear, compact, and structured latent trajectories, which well capture oscillatory patterns in neural responses for the grating stimulus. Meanwhile, the proposed model is able to convey interpretable sinusoidal stimulus patterns in $2$D rings without including the external stimulus as the model variable. Therefore, GP-RNN with nonlinear dynamics and nonlinear embedding function can help extract latent patterns more efficiently. Although P-GPLVM also achieves promising results compared with PfLDS (still worse than our GP-RNN), P-GPLVM needs much more effort than GP-RNN to fine-tune the optimization hyperparameters.


We next show the quantitative prediction performance of multiple methods. The evaluation procedure is well known as ``co-smoothing'' \cite{zhao2017variational,wu2017gaussian}, which is a standard leave-one-neuron-out test. We select all the trials with $0^\circ$, $90^\circ$, $180^\circ$, and $270^\circ$ orientations of sinusoidal grating drifting. We split all the trials into training sets (40 trials) and test sets (10 trials). The model-specific parameters, e.g., RNN dynamics and GP mapping function for GP-RNN, are estimated using training sets (all neurons). Then we fix the estimated model parameters and leave one neuron in test trials out and infer latent trajectories based on the remaining neurons. The left-out neuron spiking activity is then predicted given inferred latents of test trials and estimated parameters from training trials. Consistent with the results reported in the previous literature, the prediction is quantified by $R^{2}$. It shows the prediction performance of the firing rates compared with empirical firing rates of the left-out neurons. We iterate over all neurons as left-out ones and average the prediction $R^{2}$ values for each model shown in Table.~\ref{PredictiveR}. In this neural dataset, each recently proposed method can only increase the $R^{2}$ value by a small amount, which is still non-trivial to achieve. GP-RNN has already doubled the increment from PfLDS ($13\%$ increase of $R^{2}$ value) to P-GPLVM ($7\%$ increase).
\begin{table}[htbp!]
\centering\scalebox{0.75}{
\begin{tabular}{c|cccccc}
\toprule
Dim & PLDS & P-GPFA & LFADS & PfLDS & P-GPLVM &  GP-RNN \\\hline
2                           & 0.68 & 0.69   & 0.73  & 0.73  & 0.74    &  \underline{\textbf{0.77}}   \\\hline
4                           & 0.69 & 0.72   & 0.74  & 0.73  & 0.75    & \underline{\textbf{0.78}}   \\\hline
6                           & 0.72 & 0.73   & 0.74  & 0.74  & 0.77    &  \underline{\textbf{0.80}}   \\\hline
8                           & 0.74 & 0.74   & 0.75  & 0.75  & 0.77    &  \underline{\textbf{0.80}}   \\\hline
10                          & 0.75 & 0.74   & 0.77  & 0.76  & 0.77    &  \underline{\textbf{0.81}} \\
\bottomrule
\end{tabular}}
\caption{Predictive $R^{2}$ on neural spiking activity of test dataset. The column ``Dim'' indicates the dimension of latent process $\textbf{z}$. GP-RNN has consistently the best performance when increasing predefined latent dimensions.}
\label{PredictiveR}
\end{table}
P-GPLVM and PfLDS have comparable results and we think they benefit from nonlinear mapping functions, i.e., feed-forward neural network and Gaussian process. PLDS and P-GPFA use linear mapping but cannot capture nonlinear embeddings, and require more latent dimensionality to achieve similar results as P-GPLVM and PfLDS. Our GP-RNN with RNN dynamics and GP mapping provides the most competitive prediction accuracy, due to its nonlinear dynamical model encoding time dependence and complex nonlinear embedding function with uncertainty propagation. 

\subsection{Implementation Notes}
We have encountered the risk of over-parameterization during our experiments. When the algorithm breaks down, increasing the number of hidden nodes of RNN structures cannot improve the results much. We successfully avoid it via (1) using cross-validation to choose the number of hidden states (the risk happened with more than $30$ hidden nodes in this experimental dataset); (2) adopting Dropout(0.3)/L2 regularization for RNN gates. Too many hidden states of RNN dynamics will lead to learning both hidden states $\nu$ and cell states $c$ failure, also too few hidden states report much lower prediction performance (we fix $30$ hidden nodes ultimately in our experiments); (3) applying orthogonal initialization for RNN gates and clipping gradients tricks during training; and (4) instead of marginalizing out the latent function $f$ in the Poisson model, adopting the composing inference strategy and using GPFA to initialize $f$. The experiments are benefited from the probabilistic modeling library ``Edward''~\cite{tran2017deep}.

With respect to the stable learning process, it is robust when applying orthogonal initialization for RNN gates, Xavier Initialization for parameters of fully connected layers (mapping hidden states $\nu$ to latent states $z$), and clipping gradients tricks during training. This combination is a relatively effective way of eliminating exploding and vanishing gradients, and provides a robust learning process. 

Concerning sample perturbations, in the simulation, we randomly (both Poisson and Gaussian noise) generated the observations and parameters of mapping functions (Gaussian noise) for $10$ times; and with real neural data, we shuffled the training/testing datasets for $10$ times. The learning was based on these sample perturbations (trial variants) and the above-mentioned initialization strategies. The analysis of the sample perturbations are listed in the supplementary materials with standard errors.

\section{CONCLUSION}
To discover the insightful latent structure from neural data, we propose an unsupervised Gaussian process recurrent neural network (GP-RNN), utilizing the representation power of recurrent neural networks and the flexible nonlinear mapping function with Gaussian process. We show that GP-RNN is superior at recovering more structured latent trajectories as well as having better quantitative performance compared with other state-of-the-art methods. Besides the visual cortex dataset tested in the paper, the proposed model can also be potentially applied to analyzing the neural dynamics of primary motor cortex, prefrontal cortex (PFC) or posterior parietal cortex (PPC) which plays a significant role in cognition (evidence integration, short term memory, spatial reasoning, etc.). The model can also be applied to other domains, e.g., finance, healthcare, for extracting low-dimensional, underlying latent states from complicated time series. Our codes and additional materials are available at $\texttt{\url{https://github.com/sheqi/GP-RNN_UAI2019}}$.

\newpage

\bibliography{References}

\begin{thebibliography}{10}

\bibitem{byron2009gaussian}
M~Yu Byron, John~P Cunningham, Gopal Santhanam, Stephen~I Ryu, Krishna~V
  Shenoy, and Maneesh Sahani.
\newblock Gaussian-process factor analysis for low-dimensional single-trial
  analysis of neural population activity.
\newblock In {\em Advances in Neural Information Processing Systems (NeurIPS)},
  pages 1881--1888, 2009.

\bibitem{macke2011empirical}
Jakob~H Macke, Lars Buesing, John~P Cunningham, M~Yu Byron, Krishna~V Shenoy,
  and Maneesh Sahani.
\newblock Empirical models of spiking in neural populations.
\newblock In {\em Advances in Neural Information Processing Systems (NeurIPS)},
  pages 1350--1358, 2011.

\bibitem{gao2015high}
Yuanjun Gao, Lars Busing, Krishna~V Shenoy, and John~P Cunningham.
\newblock High-dimensional neural spike train analysis with generalized count
  linear dynamical systems.
\newblock In {\em Advances in Neural Information Processing Systems (NeurIPS)},
  pages 2044--2052, 2015.

\bibitem{gao2016linear}
Yuanjun Gao, Evan~W Archer, Liam Paninski, and John~P Cunningham.
\newblock Linear dynamical neural population models through nonlinear
  embeddings.
\newblock In {\em Advances in Neural Information Processing Systems (NeurIPS)},
  pages 163--171, 2016.

\bibitem{wu2017gaussian}
Anqi Wu, Nicholas~G Roy, Stephen Keeley, and Jonathan~W Pillow.
\newblock Gaussian process based nonlinear latent structure discovery in
  multivariate spike train data.
\newblock In {\em Advances in Neural Information Processing Systems (NeurIPS)},
  pages 3499--3508, 2017.

\bibitem{pandarinath2018inferring}
Chethan Pandarinath, Daniel~J O’Shea, Jasmine Collins, Rafal Jozefowicz,
  Sergey~D Stavisky, Jonathan~C Kao, Eric~M Trautmann, Matthew~T Kaufman,
  Stephen~I Ryu, Leigh~R Hochberg, et~al.
\newblock Inferring single-trial neural population dynamics using sequential
  auto-encoders.
\newblock {\em Nature methods}, page~1, 2018.

\bibitem{she2018reduced}
Qi~She, Yuan Gao, Kai Xu, and Rosa~HM Chan.
\newblock Reduced-rank linear dynamical systems.
\newblock In {\em Thirty-Second AAAI Conference on Artificial Intelligence
  (AAAI)}, 2018.

\bibitem{krishnan2017structured}
Rahul~G Krishnan, Uri Shalit, and David Sontag.
\newblock Structured inference networks for nonlinear state space models.
\newblock In {\em The Thirty-first AAAI Conference on Artificial Intelligence
  (AAAI)}, pages 2101--2109, 2017.

\bibitem{she2018stochastic}
Qi~She and Rosa~HM Chan.
\newblock Stochastic dynamical systems based latent structure discovery in
  high-dimensional time series.
\newblock In {\em 2018 IEEE International Conference on Acoustics, Speech and
  Signal Processing (ICASSP)}, pages 886--890. IEEE, 2018.

\bibitem{lawrence2004gaussian}
Neil~D Lawrence.
\newblock Gaussian process latent variable models for visualisation of high
  dimensional data.
\newblock In {\em Advances in Neural Information Processing Systems (NeurIPS)},
  pages 329--336, 2004.

\bibitem{nam2015poisson}
Hooram Nam.
\newblock Poisson extension of gaussian process factor analysis for modeling
  spiking neural populations.
\newblock {\em Master's thesis, Department of Neural Computation and Behaviour,
  Max Planck Institute for Biological Cybernetics, T{\"u}bingen}, 2015.

\bibitem{yu2009variance}
Guan Yu.
\newblock Variance stabilizing transformations of poisson, binomial and
  negative binomial distributions.
\newblock {\em Statistics \& Probability Letters}, 79(14):1621--1629, 2009.

\bibitem{hoffman2013stochastic}
Matthew~D Hoffman, David~M Blei, Chong Wang, and John Paisley.
\newblock Stochastic variational inference.
\newblock {\em The Journal of Machine Learning Research (JMLR)},
  14(1):1303--1347, 2013.

\bibitem{christopher2016pattern}
M~Bishop Christopher.
\newblock {\em Pattern recognition and machine learning}.
\newblock Springer-Verlag New York, 2016.

\bibitem{chung2015recurrent}
Junyoung Chung, Kyle Kastner, Laurent Dinh, Kratarth Goel, Aaron~C Courville,
  and Yoshua Bengio.
\newblock A recurrent latent variable model for sequential data.
\newblock In {\em Advances in Neural Information Processing Systems (NeurIPS)},
  pages 2980--2988, 2015.

\bibitem{gregor2015draw}
Karol Gregor, Ivo Danihelka, Alex Graves, Danilo~Jimenez Rezende, and Daan
  Wierstra.
\newblock Draw: A recurrent neural network for image generation.
\newblock {\em arXiv preprint arXiv:1502.04623}, 2015.

\bibitem{frigola2014variational}
Roger Frigola, Yutian Chen, and Carl~Edward Rasmussen.
\newblock Variational gaussian process state-space models.
\newblock In {\em Advances in Neural Information Processing Systems (NeurIPS)},
  pages 3680--3688, 2014.

\bibitem{nguyen2014collaborative}
Trung~V Nguyen, Edwin~V Bonilla, et~al.
\newblock Collaborative multi-output gaussian processes.
\newblock In {\em Annual Conference on Uncertainty in Artificial Intelligence
  (UAI)}, pages 643--652, 2014.

\bibitem{mattos2015recurrent}
C{\'e}sar Lincoln~C Mattos, Zhenwen Dai, Andreas Damianou, Jeremy Forth,
  Guilherme~A Barreto, and Neil~D Lawrence.
\newblock Recurrent gaussian processes.
\newblock {\em arXiv preprint arXiv:1511.06644}, 2015.

\bibitem{svensson2016computationally}
Andreas Svensson, Arno Solin, Simo S{\"a}rkk{\"a}, and Thomas Sch{\"o}n.
\newblock Computationally efficient bayesian learning of gaussian process state
  space models.
\newblock In {\em Artificial Intelligence and Statistics (AISTATS), 2016
  International Conference on}, pages 213--221, 2016.

\bibitem{eleftheriadis2017identification}
Stefanos Eleftheriadis, Tom Nicholson, Marc Deisenroth, and James Hensman.
\newblock Identification of gaussian process state space models.
\newblock In {\em Advances in Neural Information Processing Systems (NeurIPS)},
  pages 5309--5319, 2017.

\bibitem{ranganath2014black}
Rajesh Ranganath, Sean Gerrish, and David Blei.
\newblock Black box variational inference.
\newblock In {\em Artificial Intelligence and Statistics (AISTATS), 2014
  International Conference on}, pages 814--822, 2014.

\bibitem{archer2015black}
Evan Archer, Il~Memming Park, Lars Buesing, John Cunningham, and Liam Paninski.
\newblock Black box variational inference for state space models.
\newblock {\em arXiv preprint arXiv:1511.07367}, 2015.

\bibitem{kingma2013auto}
Diederik~P Kingma and Max Welling.
\newblock Auto-encoding variational bayes.
\newblock {\em arXiv preprint arXiv:1312.6114}, 2013.

\bibitem{damianou2016variational}
Andreas~C Damianou, Michalis~K Titsias, and Neil~D Lawrence.
\newblock Variational inference for latent variables and uncertain inputs in
  gaussian processes.
\newblock {\em The Journal of Machine Learning Research (JMLR)},
  17(1):1425--1486, 2016.

\bibitem{tran2017deep}
Dustin Tran, Matthew~D. Hoffman, Rif~A. Saurous, Eugene Brevdo, Kevin Murphy,
  and David~M. Blei.
\newblock Deep probabilistic programming.
\newblock In {\em International Conference on Learning Representations (ICLR)},
  2017.

\bibitem{zhao2017variational}
Yuan Zhao and Il~Memming Park.
\newblock Variational latent gaussian process for recovering single-trial
  dynamics from population spike trains.
\newblock {\em Neural Computation}, 29(5):1293--1316, 2017.

\bibitem{linderman2017bayesian}
Scott Linderman, Matthew Johnson, Andrew Miller, Ryan Adams, David Blei, and
  Liam Paninski.
\newblock Bayesian learning and inference in recurrent switching linear
  dynamical systems.
\newblock In {\em Artificial Intelligence and Statistics (AISTATS), 2017
  International Conference on}, pages 914--922, 2017.

\bibitem{graf2011decoding}
Arnulf~BA Graf, Adam Kohn, Mehrdad Jazayeri, and J~Anthony Movshon.
\newblock Decoding the activity of neuronal populations in macaque primary
  visual cortex.
\newblock {\em Nature neuroscience}, 14(2):239, 2011.

\end{thebibliography}
\bibliographystyle{unsrt}

\end{document}